\documentclass{article}

\usepackage[final]{neurips_2025} 

\usepackage[utf8]{inputenc}
\usepackage[T1]{fontenc}
\usepackage{hyperref}
\usepackage{url}
\usepackage{booktabs}
\usepackage{amsfonts}
\usepackage{amsmath}
\usepackage{amssymb}
\usepackage{nicefrac}
\usepackage{microtype}
\usepackage{graphicx}
\usepackage{xcolor}
\usepackage{pgfplots}
\usepackage{subcaption}
\usepackage{tikz}
\usepackage{algorithm}
\usepackage{algorithmic}
\usetikzlibrary{shapes.geometric, arrows.meta, positioning, fit, backgrounds}
\pgfplotsset{compat=1.18}

\title{Speculative Pre-Positioning:\\Decoding Stateful Sessions to the Next Decision Point Off the Critical Path}

\author{
  Victor Norgren \\
  LayerScale, Inc. \\
  \url{https://layerscale.ai} \\
  \texttt{victor@layerscale.ai}
}

\begin{document}

\maketitle

\begin{abstract}
A stateless inference server (vLLM, SGLang, TensorRT-LLM) sits idle between requests, the accelerator doing nothing while it waits. A stateful session keeps its state alive across requests, turning that idle time into a resource: between a data update and the next query when streaming, or between a tool call and its result when agentic. \emph{Speculative pre-positioning} fills those windows by decoding the session forward to its next \emph{decision point} with the target model's own forward pass and no draft model. The next request resumes from a pre-paid entry on only its delta, or, when a calibrated confidence gate fires, is answered from a cached output distribution in a single near-constant vocabulary scan with no decode at all. The cross-request prefill and entry-decode thus move off the critical path into amortized idle time, at a cost of energy and a rare, bounded false accept.

The payoff is conditional on model capability. A larger target model that reads the domain task confidently fires the gate at near-full coverage and roughly $87\%$ precision, while a smaller model never clears it, so the fast path helps only when the model is confident. When it fires, the measured first token returns in about $1.0$\,ms (P50 over $128$ fires): a prefix cache still pays the one decode step a pre-position removes (about $39$\,ms here), a nearly fortyfold reduction, and more than fiftyfold against the stateful cold path that also re-prefills the entry. We report end-to-end first-token latency with and without pre-positioning, the gate calibration, the cross-request hit rate, and the idle-time energy cost from a host run.
\end{abstract}

\section{Introduction}

Stateful inference keeps a session's state alive across requests so each one advances by only its delta, as in continuous data streaming \cite{norgren2026layerscale} and multi-agent tool calling \cite{norgren2026toolcall}. The same persistence exposes an untapped resource, the \emph{idle window}: between a data update and the next query, or between a tool call and its result, the session holds state while the accelerator idles on the critical path.

These windows are set by an external clock, not the model: a streaming query arrives on its own cadence and a tool call blocks on an external round trip, each routinely far longer than a forward pass over a short entry. The longer the window relative to that work, the larger the fraction of pre-positioned work consumed before the next state change, the hit rate of Eq.~\eqref{eq:hitrate}, which we measure rather than assume (Section~\ref{sec:measured}).

We use that window to do the next request's entry work in advance, decoding the session forward to its next \emph{decision point}, the position where generation resumes: the query prefix plus assistant header in a streaming session, or the post-tool-call envelope in an agentic one. One forward pass of the target model there, with no draft model, caches the output distribution, so the next request either resumes delta-only from a pre-paid entry or returns a single cached token when a calibrated confidence gate fires. This is not draft-model speculative decoding \cite{leviathan2023fast}: no draft network and no fast-path verification, only the target model's own work moved off the critical path, gated by a confidence test rather than a draft-versus-target match.

This paper makes the following contributions:
\begin{enumerate}
    \item \textbf{Cross-request idle-time pre-positioning, formalized.} We define the decision point, the amortization model, and the condition under which the work is wall-clock-free, costing only energy and a rare false accept.
    \item \textbf{The ready-distribution cache.} We cache the \emph{output distribution at the decision point}, not merely the KV state, enabling a zero-decode answer for an eligible query class.
    \item \textbf{A confidence-gated single-token fast path.} One cached token is served when the top-1 to top-2 logit gap clears a calibrated threshold and the token lies in a domain fast-answer set; otherwise it falls through to a normal path made cheaper by the pre-paid prefill.
    \item \textbf{Generalization across session types.} One mechanism covers streaming sessions (query prefix plus assistant header) and agentic sessions (post-tool-call envelope, the next turn resuming delta-only).
    \item \textbf{Invalidation discipline.} Any mutation of the persistent state invalidates the cached distribution and pre-positioned entry, so a stale or low-confidence answer is never served.
\end{enumerate}

The remainder formalizes the model (Section~\ref{sec:problem}), develops the mechanisms with two algorithms and supporting figures (Section~\ref{sec:arch}), and reports the methodology and results, including systems measurements from a host run (Sections~\ref{sec:setup} and \ref{sec:results}). Discussion, related work, and conclusions follow (Sections~\ref{sec:discussion}--\ref{sec:conclusion}).

\section{Problem Formulation}
\label{sec:problem}

We consider a single stateful session whose persistent KV cache survives across requests and is advanced incrementally, as in the streaming \cite{norgren2026layerscale} and multi-agent tool-calling \cite{norgren2026toolcall} settings. The session holds an accumulated state of $n_s$ tokens: a frozen system region, the streamed data or conversation history, and any committed turns. The property we exploit is temporal: the next request does not arrive the instant the state is ready. Between the moment the state reaches version $t$ (a data batch committed, or a tool call emitted) and the moment the next request arrives, the accelerator is idle. We formalize that window, the position the next request resumes from, and the latency and risk of answering it ahead of time.

\subsection{The Decision Point and the Idle Window}

When the next request arrives it does not start from scratch: it appends a short, structurally predictable \emph{entry} of $L_{\text{entry}}$ tokens to the live end of the state. In a streaming session the entry is the query prefix together with the assistant header; in an agentic session it is the post-tool-call envelope that precedes the next turn. The \emph{decision point} is the position at which generation resumes:
\begin{equation}
p \;=\; n_s + L_{\text{entry}}
\label{eq:decision-point}
\end{equation}
The model's output distribution at position $p$, the logits produced by a forward pass whose last token is the final entry token, is the \emph{ready distribution}: the model's answer to the next request, formed before that request has been asked. For the streaming fast path the pre-position runs at the data decision point on the generic assistant header, before the specific query text is known, so the ready distribution is a function of the data state and the header but not of the query: the served fast-path token is \emph{question-independent}. In Eq.~\eqref{eq:decision-point} the fast path therefore takes $L_{\text{entry}}$ as the header alone, with the query prefix folded into the fall-through delta $\Delta$ of Eq.~\eqref{eq:fallthrough}. A query whose answer is a property of the data state, the trend of the streamed series, matches this token; one whose answer turns on the query wording need not, and that gap is the false-accept rate $r(\tau)$ of Eq.~\eqref{eq:risk} measured in Section~\ref{sec:results}.

Let the state reach version $t$ at time $0$ and the next request arrive at time $I_{\text{idle}}$. We call $I_{\text{idle}}$ the \emph{idle window}. It is genuinely idle when no other admitted work occupies the accelerator over $[0, I_{\text{idle}})$, the common case for a session between a data update and a sporadic query, or between a dispatched tool call and a returning result.

\subsection{Critical-Path Latency: Baseline, Fast Path, and Fall-Through}

We reuse the latency notation of the companions: $T_{\text{prefill}}(k)$ is the time to process $k$ tokens, $T_{\text{decode}}$ the time for one generated token, $T_{\text{restore}}$ a constant-time metadata restore of cached state, and $m$ the number of output tokens. On the stateful-session baseline the accumulated state is already cached, but the entry is processed on the critical path when the request arrives, followed by the decode that produces the first answer token:
\begin{equation}
L_{\text{cold}} \;=\; T_{\text{prefill}}(L_{\text{entry}}) + T_{\text{decode}}
\label{eq:lcold}
\end{equation}
plus $(m-1)\,T_{\text{decode}}$ of tail generation for multi-token answers, which pre-positioning does not change and we omit from the entry-latency comparison.

Speculative pre-positioning runs the forward pass of Eq.~\eqref{eq:lcold} during the idle window instead, caches the ready distribution, and on an eligible request serves a single token by scanning that distribution for its argmax. The only critical-path cost is that scan over the vocabulary $\mathcal{V}$:
\begin{equation}
L_{\text{fast}} \;=\; T_{\text{scan}}(\mathcal{V})
\label{eq:lfast}
\end{equation}
which is near-constant and independent of $L_{\text{entry}}$, $n_s$, and the accumulated context. The fast-path speedup is
\begin{equation}
S \;=\; \frac{L_{\text{cold}}}{L_{\text{fast}}} \;=\; \frac{T_{\text{prefill}}(L_{\text{entry}}) + T_{\text{decode}}}{T_{\text{scan}}(\mathcal{V})}
\label{eq:fastspeedup}
\end{equation}
and grows with the entry length, since $L_{\text{fast}}$ does not.

When the gate of Section~\ref{sec:gate-formulation} does not fire, the request falls through to the normal path, but does not re-process the whole entry: the entry prefill was already committed to the cache during the idle window, so the request restores it in constant time and processes only the delta $\Delta$ it adds beyond the pre-positioned entry:
\begin{equation}
L_{\text{fall}} \;=\; T_{\text{restore}} + T_{\text{prefill}}(\Delta) + T_{\text{decode}}, \qquad \Delta \le L_{\text{entry}}
\label{eq:fallthrough}
\end{equation}
Comparing with the baseline, $L_{\text{fall}} - L_{\text{cold}} = T_{\text{restore}} - T_{\text{prefill}}(L_{\text{entry}} - \Delta)$: the fall-through trades re-prefilling the pre-paid portion of the entry for one state restore, and improves on the baseline once that saved prefill exceeds the restore cost. With $\Delta = 0$ it is $T_{\text{restore}} + T_{\text{decode}}$, and at the upper bound $\Delta = L_{\text{entry}}$ it is $L_{\text{cold}} + T_{\text{restore}}$, the restore overhead with no prefill saved. The fall-through is thus bounded within a restore of the baseline and is strictly cheaper for a pre-positioned entry longer than the restore cost, independent of whether the gate fires.

\subsection{Idle-Window Amortization, Hit Rate, and Net Benefit}

The pre-position cost is one forward pass over the entry tokens, whose final-position logits are the ready distribution:
\begin{equation}
T_{\text{pp}} \;=\; T_{\text{prefill}}(L_{\text{entry}})
\label{eq:ppcost}
\end{equation}
This work is \emph{wall-clock-free} precisely when it fits inside the idle window:
\begin{equation}
T_{\text{pp}} \;\le\; I_{\text{idle}}
\label{eq:amortization}
\end{equation}
Under Eq.~\eqref{eq:amortization} the forward pass occupies otherwise-idle accelerator time and adds nothing to query latency; the only real costs are energy and the rare false accept of Section~\ref{sec:gate-formulation}. The cached ready distribution adds negligible state beyond the session's $M_{\text{KV}}$: a single vocabulary-sized logit vector, or, in the gated fast path, only the top token and the logit gap.

Not every pre-position is consumed. If the session state mutates before the next request (a new data batch, a rollback), the cached distribution is invalidated (Section~\ref{sec:invalidation}) and the pre-position was wasted. Let $h$ be the \emph{hit rate}, the fraction of pre-positions consumed by a request before invalidation; the \emph{wasted-work fraction} is then
\begin{equation}
w \;=\; 1 - h
\label{eq:hitrate}
\end{equation}
Writing $g$ for the probability the gate fires given a hit, the expected entry latency under pre-positioning is
\begin{equation}
\mathbb{E}[L_{\text{pp}}] \;=\; h\big(g\,L_{\text{fast}} + (1-g)\,L_{\text{fall}}\big) + (1-h)\,L_{\text{cold}}
\label{eq:expected-latency}
\end{equation}
against a baseline of $\mathbb{E}[L_{\text{base}}] = L_{\text{cold}}$. Pre-positioning lowers expected latency, $\mathbb{E}[L_{\text{pp}}] < L_{\text{cold}}$, exactly when
\begin{equation}
g\,L_{\text{fast}} + (1-g)\,L_{\text{fall}} \;<\; L_{\text{cold}}
\label{eq:netbenefit}
\end{equation}
Whenever this served term falls below $L_{\text{cold}}$, expected latency drops for any hit rate $h > 0$. It does so at the measured operating point: the fast path $L_{\text{fast}}$ of Eq.~\eqref{eq:lfast} is far below $L_{\text{cold}}$ and dominates the term at an appreciable gate-fire rate $g$, and a typical fall-through of Eq.~\eqref{eq:fallthrough} is itself below the baseline. The wasted fraction $w$ never appears as wall-clock cost under Eq.~\eqref{eq:amortization}; it bears only on the energy budget, treated separately in Section~\ref{sec:discussion}.

\subsection{The Confidence Gate and Its Risk-Coverage Tradeoff}
\label{sec:gate-formulation}

The fast path of Eq.~\eqref{eq:lfast} serves a cached token with no verification pass, so its safety rests entirely on a calibrated confidence test. Let $\ell^{(1)}$ and $\ell^{(2)}$ be the largest and second-largest logits of the ready distribution, with gap
\begin{equation}
\gamma \;=\; \ell^{(1)} - \ell^{(2)}
\label{eq:gap}
\end{equation}
and let $\mathcal{A}$ be a domain \emph{fast-answer set} of admissible single-token answers. The gate fires, and a single cached token is served, iff
\begin{equation}
\gamma > \tau \;\;\text{and}\;\; \arg\max(\ell) \in \mathcal{A}
\label{eq:gate}
\end{equation}
for a threshold $\tau$ (the reference uses $\tau = 2.0$, inherited from the streaming companion's ready-position exit \cite{norgren2026layerscale}). Sweeping $\tau$ traces a selective-prediction tradeoff. Define the \emph{coverage} as the fraction of requests the gate accepts, and the \emph{false-accept rate} (the risk) as the fraction of served fast-path answers that disagree with the model's own full-path output:
\begin{align}
c(\tau) &= \Pr\!\big(\gamma > \tau \,\wedge\, \arg\max(\ell) \in \mathcal{A}\big) \label{eq:coverage} \\
r(\tau) &= \Pr\!\big(\text{served token wrong} \,\mid\, \text{accepted}\big) \label{eq:risk}
\end{align}
Both are monotone non-increasing in $\tau$: a higher threshold accepts fewer requests but errs less often. This is the selective-prediction formulation of El-Yaniv and Wiener \cite{elyaniv2010selective}, in which an abstaining predictor trades coverage for bounded risk; here the model abstains from the fast path by falling through to the normal decode. Calibration, deferred to Section~\ref{sec:setup} as a measurement, selects the smallest $\tau$ such that $r(\tau) \le r_{\max}$ for a target risk budget $r_{\max}$, maximizing coverage at bounded risk.

\section{Architecture}
\label{sec:arch}

The architecture realizes the thesis through a small set of cooperating mechanisms that share one goal: move the next request's decode entry point off the critical path and into the idle window, without ever serving a stale answer. Figure~\ref{fig:timeline} contrasts the baseline critical path with the pre-positioned one, where the entry work has moved into the idle window. Two mechanisms build the pre-positioned state, one for streaming sessions (Section~\ref{sec:preposition}) and one for agentic sessions (Section~\ref{sec:agentic}); a ready-distribution cache (Section~\ref{sec:ready-cache}) holds the result; a confidence-gated fast path (Section~\ref{sec:fastpath}) consumes it; and an invalidation rule (Section~\ref{sec:invalidation}) keeps it honest. Each rests on the persistent session state of the stateful-session foundation \cite{norgren2026layerscale,norgren2026toolcall}: none is possible in a stateless request loop, where the cache is discarded between requests and there is no live position at which to pre-position anything.

\subsection{Decision Points and Idle Windows}
\label{sec:decision-points}
A session exposes a live end position that advances as it ingests data or completes turns. The decision point of Eq.~\eqref{eq:decision-point} is the position from which the next request resumes generation, differing across session types only in what constitutes the entry. In a streaming session the entry is the query prefix plus the assistant header, so the decision point sits immediately after the header, where the model is about to answer the next query. In an agentic session the entry is the envelope that follows a tool call (the tool-result framing and the opening of the next assistant turn), so the decision point sits where that turn begins.

The idle window opens when the session commits its state, as a data batch finishes ingesting or a tool call is dispatched to an external service, and closes when the next request arrives. In both workloads this interval is typically long relative to a single forward pass: streaming data arrives on a cadence of seconds while queries are sporadic, and a dispatched tool call returns only after a network round trip or an external computation. Pre-positioning runs at the lowest scheduling priority and yields immediately to any arriving request, occupying the window only while it would otherwise be idle. When the window is genuinely idle, the amortization condition of Eq.~\eqref{eq:amortization} holds and the pre-position work is wall-clock-free.

\subsection{Pre-Positioning on Idle}
\label{sec:preposition}
The streaming mechanism produces the pre-positioned state after each update that advances the session. The engine appends the entry tokens at the live end position and runs a single forward pass (Algorithm~\ref{alg:preposition}); the logits at the final entry position are the ready distribution of Section~\ref{sec:problem}, the model's answer to the anticipated query, computed before the query is asked. It records three fields on the session: the decision-point position, the cached ready distribution, and a validity flag that marks the entry decoded and the distribution current.

Two properties make this safe and cheap. First, the cost is a forward pass over the entry tokens alone, $T_{\text{pp}} = T_{\text{prefill}}(L_{\text{entry}})$ of Eq.~\eqref{eq:ppcost}, not over the accumulated context, which already resides in the session's cache. Second, the work is best-effort: if admitting the entry tokens would breach the session's share of the cell budget, for instance under memory pressure from concurrent tenants, the engine skips the pre-position rather than evicting live state, and the next request takes the baseline path of Eq.~\eqref{eq:lcold}. Pre-positioning therefore never reduces correctness or capacity; at worst it does nothing.

\begin{algorithm}[htbp]
\caption{Pre-Position on Idle (Background, after a session update)}
\label{alg:preposition}
\begin{algorithmic}[1]
\REQUIRE Session at end position $n_s$, entry tokens $E$ of length $L_{\text{entry}}$, cell-budget headroom $B$
\STATE \COMMENT{Triggered after a data update or append; runs at lowest priority, yields to any request}
\IF{$L_{\text{entry}} > B$}
    \STATE $\textit{valid} \leftarrow \textbf{false}$ \COMMENT{Cell budget over-committed: skip without eviction; baseline path $L_{\text{cold}}$ applies}
\ELSE
    \STATE $\text{Append}(E)$ at end position $n_s$ \COMMENT{Entry only; accumulated state already resident}
    \STATE $\ell_{\text{ready}} \leftarrow \text{Forward}(E)$ \COMMENT{One forward pass, cost $T_{\text{pp}}$}
    \STATE $p \leftarrow n_s + L_{\text{entry}}$ \COMMENT{Decision point}
    \STATE
    \STATE \COMMENT{Record the ready distribution on the session}
    \STATE $\textit{position} \leftarrow p$
    \STATE $\textit{dist} \leftarrow \ell_{\text{ready}}$ \COMMENT{Or only $(\arg\max \ell_{\text{ready}},\, \gamma)$ under the gate}
    \STATE $\textit{decoded} \leftarrow \textbf{true};\;\; \textit{valid} \leftarrow \textbf{true}$
\ENDIF
\end{algorithmic}
\end{algorithm}

\subsection{The Ready-Distribution Cache}
\label{sec:ready-cache}
The pre-positioned forward pass leaves two kinds of state behind, and the distinction is the crux of the contribution. The first is the set of KV cells for the entry tokens, which persist in the session's cache as a prefix cache would retain them; a request that falls through resumes from them in constant time, the $T_{\text{restore}}$ term of Eq.~\eqref{eq:fallthrough}. The second, which conventional caches do not keep, is the output distribution at the decision point: the ready logits, or in the common gated case only the top token and the top-1 to top-2 gap. A prefix or KV cache \cite{kwon2023vllm,zheng2024sglang} stores the first kind alone, which spares a request the entry prefill but still leaves it to run the entry decode to discover what the model would say. Caching the distribution lets an eligible request skip the decode as well and answer directly from the cached argmax, collapsing the critical path to the vocabulary scan $L_{\text{fast}}$ of Eq.~\eqref{eq:lfast}. That distribution is shared with the registered-query path of the streaming companion \cite{norgren2026layerscale}, so a standing query and an ad-hoc query resolving to the same decision point draw the same pre-computed answer. It lives on the session beside the KV state and adds negligible memory: at most one vocabulary-sized vector against the session's $M_{\text{KV}}$.

\subsection{Confidence-Gated Fast-Path Serve}
\label{sec:fastpath}
A request is \emph{eligible} for the fast path iff the session holds a valid ready distribution and the request's realized entry reproduces the pre-positioned entry token-for-token up to the decision point of Eq.~\eqref{eq:decision-point}: only then is the cached argmax provably identical to a fresh forward pass on the live sequence (Section~\ref{sec:invalidation}), so an exact-match request, such as an agentic delta-only resume or a registered query, takes on zero added risk. The streaming ad-hoc fast path relaxes the exact-match condition, serving the question-independent token formed at the header before the query arrives (Section~\ref{sec:problem}) and accepting the bounded divergence $r(\tau)$ of Eq.~\eqref{eq:risk} in exchange for latency. An eligible request takes the fast path of Eq.~\eqref{eq:lfast}; the serve procedure (Algorithm~\ref{alg:fastpath}) reads the cached distribution, forms the gap $\gamma$ of Eq.~\eqref{eq:gap} from its two largest logits, and applies the gate of Eq.~\eqref{eq:gate}: if $\gamma$ exceeds the threshold $\tau$ and the predicted token lies in the fast-answer set $\mathcal{A}$, it returns that single token and bypasses the scheduler, so the request never enters the decode loop and its whole critical path is the vocabulary scan $L_{\text{fast}}$, with no prefill and no decode. When the gate does not fire, because the gap is too small or the predicted token is not an admissible single-token answer, the request falls through to the normal path, still paying only $L_{\text{fall}}$ of Eq.~\eqref{eq:fallthrough} rather than the baseline cost, since the entry prefill was already committed during the idle window.

Running no verification is the structural difference from token-level speculation. Speculative decoding proposes tokens and checks each against the target's logits within one generation \cite{leviathan2023fast}, and prompt-lookup decoding does the same without a draft model but still verifies every proposed token \cite{saxena2023pld}. Here the served token is the target model's own output, already produced by the pre-position pass, so there is no draft to verify it against. There is, however, something it can disagree with: the streaming pre-position conditions on the data state and the generic header but not on the specific query, so the served token is question-independent, whereas the full path conditions on the query as well. When the answer is a property of the data state the two coincide; when it is not they diverge, at the measured false-accept rate $r(\tau)$ of Eq.~\eqref{eq:risk}. The fast path thus trades question-conditioning for latency. The gate is therefore a calibrated confidence test (Section~\ref{sec:gate-formulation}) whose threshold holds the false-accept rate of Eq.~\eqref{eq:risk} below a target budget, not a draft-versus-target match. The same cached distribution feeds the registered-query path of the streaming companion \cite{norgren2026layerscale}, so a standing query and an ad-hoc request arriving at the same decision point read one pre-computed confidence rather than each recomputing it.

\begin{algorithm}[htbp]
\caption{Confidence-Gated Fast-Path Serve (on request)}
\label{alg:fastpath}
\begin{algorithmic}[1]
\REQUIRE Request $q$ on a session; cached $(p,\, \ell_{\text{ready}},\, \textit{valid})$; threshold $\tau$, fast-answer set $\mathcal{A}$
\IF{$\neg\,\textit{valid}$ \textbf{or} $q$ ineligible (realized entry $\neq$ pre-positioned entry up to $p$)}
    \STATE \textbf{fall through} to the scheduler \COMMENT{Normal path, cost $L_{\text{fall}}$; entry prefill already paid}
\ELSE
    \STATE $(\ell^{(1)}, \ell^{(2)}) \leftarrow \text{Top2}(\ell_{\text{ready}})$ \COMMENT{Two largest logits of the cached distribution}
    \STATE $\gamma \leftarrow \ell^{(1)} - \ell^{(2)}$;\;\; $\hat{a} \leftarrow \arg\max(\ell_{\text{ready}})$ \COMMENT{Logit gap and top token}
    \IF{$\gamma > \tau$ \textbf{and} $\hat{a} \in \mathcal{A}$}
        \RETURN $\hat{a}$ \COMMENT{One cached token, bypass the scheduler; cost $L_{\text{fast}}$, no verification}
    \ELSE
        \STATE \textbf{fall through} to the scheduler \COMMENT{Gate not met; normal path, cost $L_{\text{fall}}$}
    \ENDIF
\ENDIF
\end{algorithmic}
\end{algorithm}

\begin{figure}[htbp]
\centering
\begin{subfigure}{0.48\textwidth}
\centering
\begin{tikzpicture}[
    every node/.style={font=\sffamily},
    start/.style={rectangle, draw, minimum width=1.7cm, minimum height=0.6cm, align=center, font=\sffamily\scriptsize, fill=gray!12},
    proc/.style={rectangle, draw, minimum width=1.9cm, minimum height=0.6cm, align=center, font=\sffamily\scriptsize, fill=blue!10},
    dec/.style={diamond, draw, aspect=1.7, inner sep=1pt, align=center, font=\sffamily\scriptsize, fill=gray!8},
    term/.style={rectangle, draw, minimum width=1.6cm, minimum height=0.6cm, align=center, font=\sffamily\scriptsize, fill=orange!25},
    good/.style={rectangle, draw, minimum width=1.7cm, minimum height=0.6cm, align=center, font=\sffamily\scriptsize, fill=blue!20},
    arrow/.style={-{Stealth[length=2mm]}, thick},
    elab/.style={font=\sffamily\scriptsize, gray}
]
\node[start] (u) at (0,0) {Session update};
\node[dec] (b) at (0,-1.7) {$L_{\text{entry}}$\\$> B$?};
\node[term] (skip) at (2.5,-1.7) {skip;\\$L_{\text{cold}}$};
\node[proc] (f) at (0,-3.4) {append;\\forward ($T_{\text{pp}}$)};
\node[proc] (r) at (0,-4.8) {store $p$,\\dist; valid};
\node[good] (v) at (0,-6.2) {pre-positioned\\+ valid};
\draw[arrow] (u) -- (b);
\draw[arrow] (b) -- node[elab, above] {yes} (skip);
\draw[arrow] (b) -- node[elab, right] {no} (f);
\draw[arrow] (f) -- (r);
\draw[arrow] (r) -- (v);
\end{tikzpicture}
\caption{Pre-position on idle (Algorithm~\ref{alg:preposition}).}
\end{subfigure}
\hfill
\begin{subfigure}{0.48\textwidth}
\centering
\begin{tikzpicture}[
    every node/.style={font=\sffamily},
    start/.style={rectangle, draw, minimum width=1.7cm, minimum height=0.6cm, align=center, font=\sffamily\scriptsize, fill=gray!12},
    dec/.style={diamond, draw, aspect=1.7, inner sep=1pt, align=center, font=\sffamily\scriptsize, fill=gray!8},
    term/.style={rectangle, draw, minimum width=1.6cm, minimum height=0.6cm, align=center, font=\sffamily\scriptsize, fill=orange!25},
    good/.style={rectangle, draw, minimum width=1.7cm, minimum height=0.6cm, align=center, font=\sffamily\scriptsize, fill=blue!20},
    arrow/.style={-{Stealth[length=2mm]}, thick},
    elab/.style={font=\sffamily\scriptsize, gray}
]
\node[start] (q) at (0,0) {request};
\node[dec] (e) at (0,-1.7) {$\textit{valid}$,\\eligible?};
\node[term] (fa) at (2.5,-1.7) {fall through\\$L_{\text{fall}}$};
\node[dec] (g) at (0,-3.7) {$\gamma{>}\tau$\\$\wedge\,\hat{a}{\in}\mathcal{A}$?};
\node[term] (fb) at (2.5,-3.7) {fall through\\$L_{\text{fall}}$};
\node[good] (serve) at (0,-5.6) {serve $\hat{a}$\\$L_{\text{fast}}$};
\draw[arrow] (q) -- (e);
\draw[arrow] (e) -- node[elab, above] {no} (fa);
\draw[arrow] (e) -- node[elab, right] {yes} (g);
\draw[arrow] (g) -- node[elab, above] {no} (fb);
\draw[arrow] (g) -- node[elab, right] {yes} (serve);
\end{tikzpicture}
\caption{Confidence-gated serve (Algorithm~\ref{alg:fastpath}).}
\end{subfigure}
\caption{Control flow of the two pre-positioning algorithms. (a) The idle-window pre-position of Algorithm~\ref{alg:preposition}: after a session update, if admitting the entry would breach the cell-budget headroom $B$ the engine skips and the next request takes the baseline $L_{\text{cold}}$; otherwise it appends the entry, runs one forward pass at cost $T_{\text{pp}}$ of Eq.~\eqref{eq:ppcost}, records the decision point and ready distribution, and marks the session valid. (b) The confidence-gated serve of Algorithm~\ref{alg:fastpath}: an ineligible or invalidated request falls through at cost $L_{\text{fall}}$ of Eq.~\eqref{eq:fallthrough}; otherwise the logit gap $\gamma$ is tested against the gate of Eq.~\eqref{eq:gate}, serving one cached token at cost $L_{\text{fast}}$ of Eq.~\eqref{eq:lfast} when it fires and falling through to $L_{\text{fall}}$ when it does not. Diamonds are decisions, orange terminals are fall-back paths, and the blue terminal is the pre-positioned success state.}
\label{fig:algflow}
\end{figure}
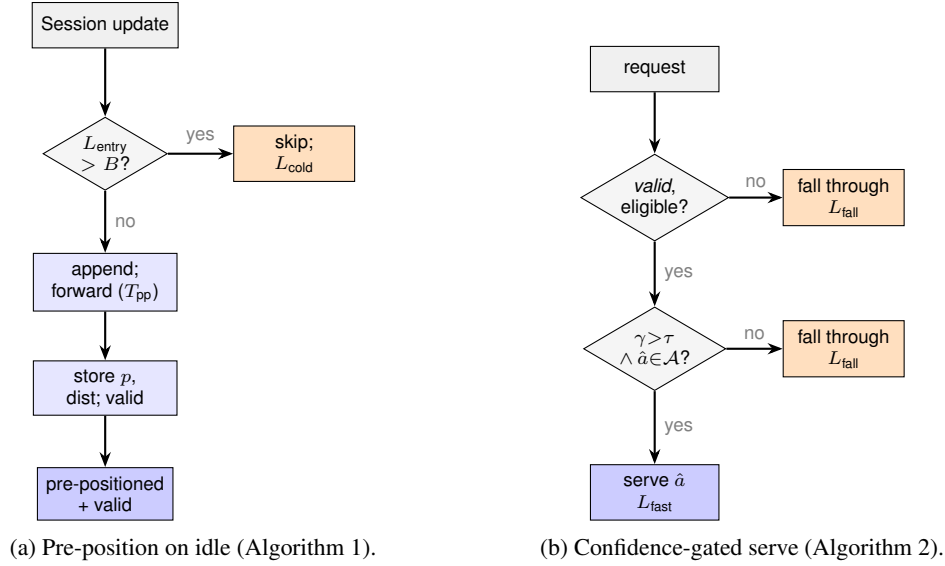

Figure~\ref{fig:algflow} renders both algorithms as flowcharts, the idle-window pre-position of Algorithm~\ref{alg:preposition} (panel a) beside the gated serve of Algorithm~\ref{alg:fastpath} (panel b), tracing each path to its cost: $T_{\text{pp}}$ off the critical path, and $L_{\text{fast}}$ or $L_{\text{fall}}$ on it.

\subsection{Generalization to Agentic Sessions}
\label{sec:agentic}
The same construction extends across the tool-call boundary, a second species of idle window. When a session emits a tool call it waits for the result, idle on the critical path for the duration of a network round trip or an external computation. During that wait the engine pre-decodes the post-tool-call envelope, the tokens that frame the returning result and open the next assistant turn, placing the decision point at the start of that turn. When the result arrives, the next request appends only its delta, the result payload, beyond the pre-positioned envelope and resumes delta-only, paying $L_{\text{fall}}$ of Eq.~\eqref{eq:fallthrough} rather than re-processing the envelope on the critical path. The entry is an envelope rather than a query prefix, but the decision-point position, the ready distribution, and the validity flag are recorded exactly as in Section~\ref{sec:preposition}, so the ready-distribution cache of Section~\ref{sec:ready-cache} and the fast path of Section~\ref{sec:fastpath} carry over unchanged; the dominant benefit here is the delta-only resume, with the single-token serve applying whenever a tool-mediated turn admits an eligible cached answer. The agentic pre-position is best-effort like the streaming one: if pre-decoding the envelope would over-commit the session's share of the cell budget, as when many agents run concurrently, the engine skips it and the next turn takes the baseline path of Eq.~\eqref{eq:lcold}, disturbing no live state. One mechanism therefore serves both the streaming analytical-query workload \cite{norgren2026layerscale} and the multi-agent tool-calling workload \cite{norgren2026toolcall}.

\subsection{Invalidation Discipline}
\label{sec:invalidation}
\begin{figure}[htbp]
\centering
\begin{tikzpicture}[
    every node/.style={font=\sffamily},
    statebox/.style={rectangle, draw, minimum width=2.4cm, minimum height=0.95cm, align=center, font=\sffamily\scriptsize},
    nostate/.style={statebox, fill=gray!12},
    validstate/.style={statebox, fill=blue!15},
    gate/.style={diamond, draw, aspect=2, inner sep=1pt, align=center, font=\sffamily\scriptsize, fill=gray!8},
    termbox/.style={rectangle, draw, minimum width=2.5cm, minimum height=0.85cm, align=center, font=\sffamily\scriptsize, fill=orange!25},
    arrow/.style={-{Stealth[length=2mm]}, thick},
    elabelg/.style={font=\sffamily\scriptsize, gray}
]

\node[nostate] (s1) at (-2.8, 0) {NO READY\\DIST};
\node[validstate] (s2) at (1.0, 0) {PRE-POSITIONED\\+ VALID};

\node[gate] (g) at (1.0, -2.4) {$\gamma > \tau$\\$\wedge\;\hat{a} \in \mathcal{A}$?};

\node[termbox] (serve) at (-1.4, -4.4) {SERVE one\\cached token ($L_{\text{fast}}$)};
\node[termbox] (fall) at (3.4, -4.4) {FALL THROUGH\\($L_{\text{fall}}$)};

\draw[arrow] ([yshift=8pt]s1.east) -- ([yshift=8pt]s2.west);
\draw[arrow] ([yshift=-8pt]s2.west) -- ([yshift=-8pt]s1.east);
\draw[arrow] (s2.south) -- (g.north);
\draw[arrow] (g) -- (serve);
\draw[arrow] (g) -- (fall);

\node[elabelg, align=center] at (-0.9, 0.95) {update / append:\\pre-position};
\node[elabelg, align=center] at (-0.9, -1.2) {state mutation: shift,\\rollback, fresh update\\(clears the flags)};
\node[elabelg, anchor=west] at (1.2, -1.45) {request};
\node[elabelg] at (-1.1, -3.15) {gate met};
\node[elabelg] at (3.2, -3.15) {gate not met};

\end{tikzpicture}
\caption{The cached-distribution lifecycle as a session state machine. An idle-window pre-position (Algorithm~\ref{alg:preposition}) carries the session from \textsc{no ready dist} (its initial state, and the state after any invalidation) to \textsc{pre-positioned + valid}, recording the decision-point position $p$ of Eq.~\eqref{eq:decision-point}, the cached distribution, and the validity flag. An arriving request consults the confidence gate of Eq.~\eqref{eq:gate}: if the logit gap $\gamma$ clears the threshold $\tau$ and the predicted token $\hat{a}$ lies in the fast-answer set $\mathcal{A}$, one cached token is served at cost $L_{\text{fast}}$; otherwise the request falls through at cost $L_{\text{fall}}$, the entry prefill already paid. Any mutation of the persistent state (a sliding-window shift, a rollback, or a fresh update past the entry) clears the flags and returns the session to \textsc{no ready dist}: the invalidation invariant of Section~\ref{sec:invalidation} that ensures a stale distribution is never served.}
\label{fig:state}
\end{figure}
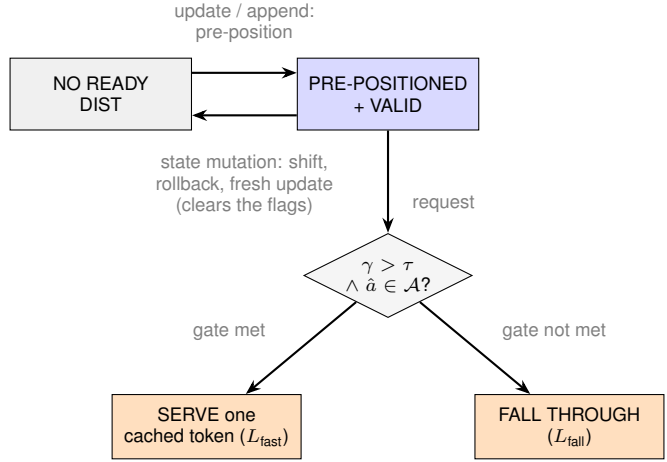

Pre-positioning is safe only if a cached distribution is never served after the state it summarizes has changed. Figure~\ref{fig:state} traces the lifecycle: an idle-window pre-position carries the session from holding no ready distribution to holding a valid one, an arriving request consults the confidence gate and is either served a cached token or falls through, and any mutation of the persistent state returns the session to the no-ready-distribution state. One invariant enforces safety: the validity flag is true only while the token sequence up to the decision point of Eq.~\eqref{eq:decision-point} is identical to the sequence from which the ready distribution was computed. Every operation that mutates that state, a sliding-window shift that relocates positions, a rollback that discards committed tokens, or a fresh update that advances past the pre-positioned entry, clears the decoded-entry and validity flags in the same step that performs the mutation, rather than deferring the test to serve time.

Correctness follows directly. The ready distribution is a deterministic function of the token sequence up to the decision point, so while that sequence is unchanged the cached argmax equals what a fresh forward pass would produce, and serving it through Eq.~\eqref{eq:lfast} is indistinguishable from the baseline of Eq.~\eqref{eq:lcold}. This is the \emph{eligibility invariant}: the fast path serves a request only when its realized entry reproduces the pre-positioned entry token-for-token up to the decision point, the exact-match condition under which the cached argmax provably equals a fresh forward pass, so an exact-match request takes on zero added risk while any request that extends the entry differently falls through. If the sequence changes, the flag is cleared atomically with it, so the fast path of Section~\ref{sec:fastpath} observes an invalid distribution and falls through to recompute on the live state; no interleaving lets a request read a distribution the current state would not reproduce. Because invalidation is coupled to the mutation rather than polled lazily, a stale or low-confidence answer is never served. An invalidation costs only a discarded pre-position, the wasted-work fraction $w$ of Eq.~\eqref{eq:hitrate}, which bears on the energy budget but never on correctness.

\begin{figure}[htbp]
\centering
\begin{tikzpicture}[
    node distance=0.4cm and 0.8cm,
    every node/.style={font=\sffamily},
    box/.style={rectangle, draw, minimum width=1.8cm, minimum height=0.7cm, align=center, font=\sffamily\scriptsize},
    triggerbox/.style={box, fill=gray!12},
    idlebox/.style={box, fill=blue!15},
    critbox/.style={box, fill=orange!25},
    arrow/.style={-{Stealth[length=2mm]}, thick},
    timeline/.style={-{Stealth[length=2mm]}, thick, gray},
    label/.style={font=\sffamily\small\bfseries},
    timelabel/.style={font=\sffamily\scriptsize, gray}
]

\node[label] at (-3.5, 2.9) {(a) Baseline};
\draw[timeline] (-3.5, 2.3) -- (4.9, 2.3);
\node[timelabel, anchor=east] at (-3.5, 2.3) {time};

\node[triggerbox] (q1) at (-2.6, 1.3) {Query};
\node[critbox] (p1) at (-0.5, 1.3) {Prefill\\(entry)};
\node[critbox] (d1) at (1.6, 1.3) {Decode\\(entry)};
\node[critbox] (t1) at (3.7, 1.3) {First\\token};

\draw[arrow] (q1) -- (p1);
\draw[arrow] (p1) -- (d1);
\draw[arrow] (d1) -- (t1);

\draw[decorate, decoration={brace, amplitude=5pt, mirror}] (-1.4, 0.7) -- (4.6, 0.7);
\node[timelabel] at (1.6, 0.25) {$L_{\text{cold}} = T_{\text{prefill}}(L_{\text{entry}}) + T_{\text{decode}}$ on the critical path};

\draw[gray, dashed] (-3.8, -0.1) -- (5.3, -0.1);

\node[label] at (-3.5, -0.9) {(b) Pre-positioned (ours)};
\draw[timeline] (-3.5, -1.5) -- (4.9, -1.5);
\node[timelabel, anchor=east] at (-3.5, -1.5) {time};

\node[triggerbox] (u1) at (-2.6, -2.5) {Update};
\node[idlebox] (pp) at (-0.5, -2.5) {Pre-position\\(forward pass)};
\node[idlebox] (rd) at (1.6, -2.5) {Ready\\distribution};

\draw[arrow] (u1) -- (pp);
\draw[arrow] (pp) -- (rd);

\draw[decorate, decoration={brace, amplitude=5pt, mirror}] (-3.5, -3.0) -- (2.5, -3.0);
\node[timelabel] at (-0.5, -3.35) {idle window (amortized, off the critical path)};

\node[triggerbox] (q2) at (1.6, -4.05) {Query};
\node[critbox] (serve) at (3.7, -3.45) {Serve scan\\($L_{\text{fast}}$)};

\draw[arrow] (q2) -- (serve);
\draw[arrow] (rd) -- (serve);

\draw[decorate, decoration={brace, amplitude=5pt, mirror}] (2.8, -4.1) -- (4.6, -4.1);
\node[timelabel] at (3.7, -4.45) {$L_{\text{fast}}$ on the critical path};

\node[idlebox, minimum width=0.45cm, minimum height=0.32cm] at (-3.05, -4.95) {};
\node[font=\sffamily\scriptsize, anchor=west] at (-2.75, -4.95) {idle-window work};
\node[critbox, minimum width=0.45cm, minimum height=0.32cm] at (0.65, -4.95) {};
\node[font=\sffamily\scriptsize, anchor=west] at (0.95, -4.95) {critical-path work};

\end{tikzpicture}
\caption{Speculative pre-positioning moves entry work off the critical path. (a) On the baseline, the query pays the entry prefill and the entry decode when it arrives, a critical-path latency $L_{\text{cold}} = T_{\text{prefill}}(L_{\text{entry}}) + T_{\text{decode}}$. (b) With pre-positioning, a single forward pass after the update computes and caches the ready distribution during the idle window, off the critical path; the query then returns a single cached token after only a vocabulary scan of near-constant cost $L_{\text{fast}}$. Blue marks idle-window work, orange marks critical-path work.}
\label{fig:timeline}
\end{figure}

\section{Experimental Setup}
\label{sec:setup}

We evaluate speculative pre-positioning with a layered methodology that separates what is measured from what is derived in closed form, so that no derived quantity is presented as measured. All measurements run on LayerScale, a proprietary inference engine with native support for the stateful sessions this work builds on, the persistent cross-request key-value cache and the live decision-point positions of Section~\ref{sec:problem}; the streaming and tool-calling companions \cite{norgren2026layerscale,norgren2026toolcall} describe that engine, and the mechanism of this paper is implemented in its serving path. The first layer is \emph{measured constants}, taken on a single H100 with a larger 70B-class target model at 4-bit precision (TP$=1$). The prefill cost per token is $\tau_p \approx 0.843$\,ms/token (HW-1), from the slope of response time against unique, cache-defeating prompt lengths (a repeated-filler prompt reads near zero, a caching artifact, so we defeat it). The fast path is subject to a model-capability condition that this run makes explicit: the confidence gate fires only when the model reads the domain task decisively, which the 70B-class model does and a smaller 8B model (BF16) does not (Section~\ref{sec:results}).
The second layer is \emph{closed-form analysis} on top of those constants: the critical-path latency, fast-path latency, and speedup of Figure~\ref{fig:speedup} follow directly from Eqs.~\eqref{eq:lcold}, \eqref{eq:lfast}, and \eqref{eq:fastspeedup}, and the idle-window amortization and net-benefit conditions of Eqs.~\eqref{eq:amortization} and \eqref{eq:netbenefit} are algebraic consequences of the model of Section~\ref{sec:problem}, not empirical fits. The third layer is \emph{measured behavior}: the selective-prediction curve of Figure~\ref{fig:risk-coverage}, Eqs.~\eqref{eq:coverage} and \eqref{eq:risk}, is drawn from the measured (logit-gap, correctness) pairs of the host run (HW-3), and the cross-request hit rate of Figure~\ref{fig:amortization} overlays the measured per-pre-position hit rate (HW-4) on the closed-loop session-trace model of Eqs.~\eqref{eq:hitrate} and \eqref{eq:expected-latency}. The session-trace model itself remains a deterministic closed form, computed by our committed simulator with no accelerator dependence.

The systems quantities that require an accelerator are measured on the host run and collected in Section~\ref{sec:measured}: the single-step decode time, the cached-distribution vocabulary-scan cost, and the state-restore cost (HW-2); the confidence-gate calibration from real (logit-gap, correctness) pairs (HW-3); the pre-position hit rate and wasted-work fraction from a live trace (HW-4); and the headline end-to-end P50 and P99 first-token latency with and without pre-positioning (HW-5). Each curve in Section~\ref{sec:results} is parameterized by these measured values, and the gate figure is drawn from the measured pairs.

\section{Results}
\label{sec:results}

\begin{figure}[htbp]
\centering
\begin{subfigure}{0.48\textwidth}
\centering
\begin{tikzpicture}
\begin{axis}[
    width=\linewidth, height=5cm,
    xlabel={Entry length $L_{\text{entry}}$ (tokens)},
    ylabel={Critical-path latency (ms)},
    xmin=0, xmax=320, ymin=0, ymax=330,
    legend pos=north west, legend cell align=left,
    grid=major,
    tick label style={font=\scriptsize},
    label style={font=\small},
    legend style={font=\scriptsize}
]
\addplot[orange!85!black, very thick, mark=none] coordinates {(16,52.605) (24,59.347) (32,66.09) (40,72.832) (48,79.574) (56,86.317) (64,93.059) (72,99.802) (80,106.544) (88,113.286) (96,120.029) (104,126.771) (112,133.514) (120,140.256) (128,146.998) (136,153.741) (144,160.483) (152,167.226) (160,173.968) (168,180.71) (176,187.453) (184,194.195) (192,200.938) (200,207.68) (208,214.422) (216,221.165) (224,227.907) (232,234.65) (240,241.392) (248,248.134) (256,254.877) (264,261.619) (272,268.362) (280,275.104) (288,281.846) (296,288.589) (304,295.331) (312,302.074)};
\addlegendentry{$L_{\text{cold}}$ (baseline)}
\addplot[blue, very thick, dashed, mark=none] coordinates {(16,0.664) (24,0.664) (32,0.664) (40,0.664) (48,0.664) (56,0.664) (64,0.664) (72,0.664) (80,0.664) (88,0.664) (96,0.664) (104,0.664) (112,0.664) (120,0.664) (128,0.664) (136,0.664) (144,0.664) (152,0.664) (160,0.664) (168,0.664) (176,0.664) (184,0.664) (192,0.664) (200,0.664) (208,0.664) (216,0.664) (224,0.664) (232,0.664) (240,0.664) (248,0.664) (256,0.664) (264,0.664) (272,0.664) (280,0.664) (288,0.664) (296,0.664) (304,0.664) (312,0.664)};
\addlegendentry{$L_{\text{fast}}$ (fast path)}
\addplot[black, only marks, mark=*, mark size=1.6pt] coordinates {(18,53.67) (34,70.39) (70,101.3) (143,167.6) (309,315.9)};
\addlegendentry{$L_{\text{cold}}$ measured}
\addplot[gray, dashed, forget plot] coordinates {(18,0) (18,330)};
\end{axis}
\end{tikzpicture}
\caption{Critical-path latency.}
\end{subfigure}
\hfill
\begin{subfigure}{0.48\textwidth}
\centering
\begin{tikzpicture}
\begin{axis}[
    width=\linewidth, height=5cm,
    xlabel={Entry length $L_{\text{entry}}$ (tokens)},
    ylabel={Fast-path speedup $S$},
    xmin=0, xmax=320, ymin=0, ymax=480,
    grid=major,
    tick label style={font=\scriptsize},
    label style={font=\small}
]
\addplot[blue!70!black, very thick, mark=none] coordinates {(16,79.248) (24,89.405) (32,99.563) (40,109.72) (48,119.877) (56,130.034) (64,140.192) (72,150.349) (80,160.506) (88,170.663) (96,180.821) (104,190.978) (112,201.135) (120,211.293) (128,221.45) (136,231.607) (144,241.764) (152,251.922) (160,262.079) (168,272.236) (176,282.393) (184,292.551) (192,302.708) (200,312.865) (208,323.023) (216,333.18) (224,343.337) (232,353.494) (240,363.652) (248,373.809) (256,383.966) (264,394.124) (272,404.281) (280,414.438) (288,424.595) (296,434.753) (304,444.91) (312,455.067)};
\end{axis}
\end{tikzpicture}
\caption{Fast-path speedup $S$.}
\end{subfigure}
\caption{Closed-form critical-path latency and fast-path speedup versus entry length, from the model of Section~\ref{sec:problem}. (a) The baseline latency $L_{\text{cold}} = T_{\text{prefill}}(L_{\text{entry}}) + T_{\text{decode}}$ of Eq.~\eqref{eq:lcold} grows linearly with the entry length, while the fast-path latency $L_{\text{fast}} = T_{\text{scan}}(\mathcal{V})$ of Eq.~\eqref{eq:lfast} is near-constant. (b) The fast-path speedup $S = L_{\text{cold}}/L_{\text{fast}}$ of Eq.~\eqref{eq:fastspeedup} therefore increases with the entry length. Curves are computed from the measured prefill cost $\tau_p \approx 0.843$\,ms/token (HW-1) with the measured decode time and the derived marginal vocabulary-scan cost (HW-2); the fast-path latency plotted here is the marginal scan, so the realized end-to-end speedup is the measured figure of Section~\ref{sec:measured}. The black markers in (a) are the measured cold first token at genuine uncached entry lengths (HW-5); they track the closed-form line across the measured span, at a slightly steeper measured slope ($\approx 0.90$\,ms/token) consistent with a small per-token overhead. The dashed vertical marks the representative measured entry of $\approx 18$ tokens, so the curve to its right is closed-form extrapolation beyond the measured operating point. Coordinates are emitted by our committed simulator.}
\label{fig:speedup}
\end{figure}
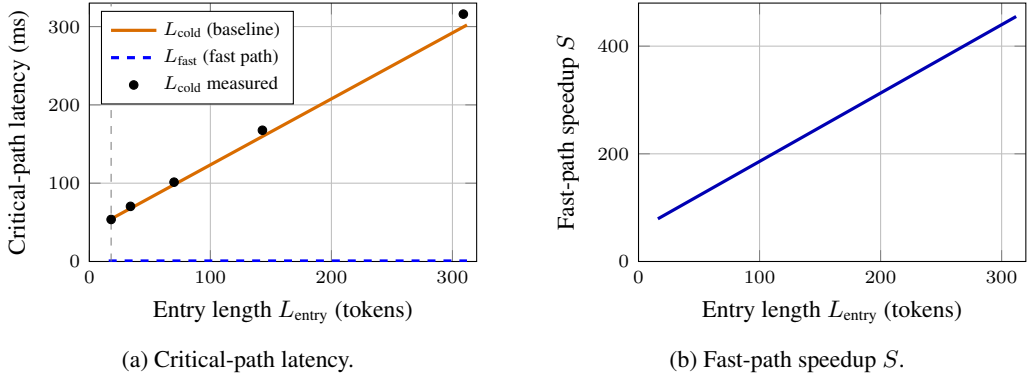

We report results in the layered structure of Section~\ref{sec:setup}: closed-form analysis parameterized by measured constants here, the measured selective-prediction and hit-rate behavior next, and the end-to-end systems measurements of Section~\ref{sec:measured}. Figure~\ref{fig:speedup} plots the closed-form critical-path latency and the resulting fast-path speedup as the entry length grows, from Eqs.~\eqref{eq:lcold}, \eqref{eq:lfast}, and \eqref{eq:fastspeedup} parameterized by the measured prefill, decode, and scan constants (HW-1, HW-2). The baseline $L_{\text{cold}}$ rises with the entry length while $L_{\text{fast}}$ stays near-constant at the marginal vocabulary scan, so the analytical speedup $S$ grows with the entry length: because $L_{\text{fast}}$ is independent of the accumulated context, the advantage is largest exactly where the baseline is most expensive. The curve uses the marginal scan cost ($\approx 0.664$\,ms); the measured end-to-end fast path adds a fixed per-request handling floor common to all served paths, so the realized first token is $1.01$\,ms (P50 over $128$ fires, HW-5) against a cold first token of $53.1$\,ms, a measured end-to-end speedup of more than fiftyfold (Section~\ref{sec:measured}).

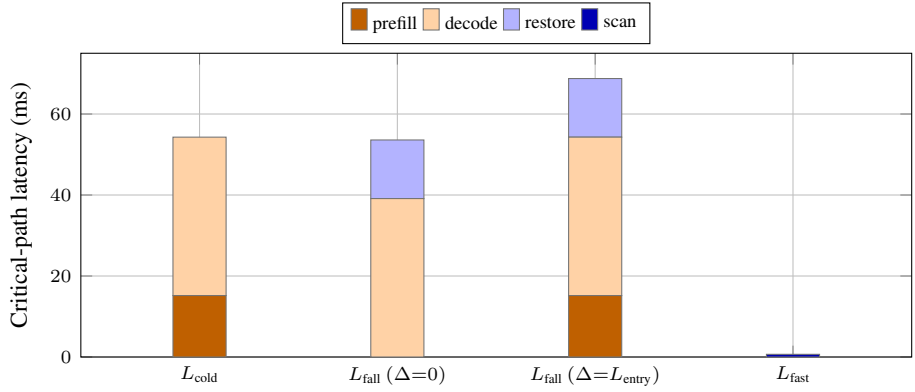
\begin{figure}[htbp]
\centering
\begin{tikzpicture}
\begin{axis}[
    width=0.9\linewidth, height=5.6cm,
    ybar stacked,
    bar width=20pt,
    ymin=0, ymax=75,
    ylabel={Critical-path latency (ms)},
    xtick={1,2,3,4},
    xticklabels={$L_{\text{cold}}$, $L_{\text{fall}}\,(\Delta{=}0)$, $L_{\text{fall}}\,(\Delta{=}L_{\text{entry}})$, $L_{\text{fast}}$},
    xmin=0.4, xmax=4.6,
    legend style={at={(0.5,1.03)}, anchor=south, legend columns=4, font=\scriptsize},
    legend cell align=left,
    tick label style={font=\scriptsize},
    label style={font=\small},
    grid=major
]
\addplot[fill=orange!75!black, draw=black!55] coordinates {(1,15.17) (2,0) (3,15.17) (4,0)};
\addlegendentry{prefill}
\addplot[fill=orange!35, draw=black!55] coordinates {(1,39.12) (2,39.12) (3,39.12) (4,0)};
\addlegendentry{decode}
\addplot[fill=blue!30, draw=black!55] coordinates {(1,0) (2,14.47) (3,14.47) (4,0)};
\addlegendentry{restore}
\addplot[fill=blue!70!black, draw=black!55] coordinates {(1,0) (2,0) (3,0) (4,0.664)};
\addlegendentry{scan}
\end{axis}
\end{tikzpicture}
\caption{Cost anatomy of the three served paths at the measured representative entry length of $18$ tokens, each decomposed into additive components from the model of Section~\ref{sec:problem}. The baseline $L_{\text{cold}}$ of Eq.~\eqref{eq:lcold} is the entry prefill plus one decode; the fall-through $L_{\text{fall}}$ of Eq.~\eqref{eq:fallthrough} is a constant-time restore plus one decode at the exact-match endpoint $\Delta = 0$ and adds a full residual prefill at the upper-bound endpoint $\Delta = L_{\text{entry}}$; the fast path $L_{\text{fast}}$ of Eq.~\eqref{eq:lfast} is the lone vocabulary scan. At this short measured entry the single decode step dominates every served path and the entry prefill is small, so $L_{\text{fall}}(\Delta{=}0)$ nearly coincides with $L_{\text{cold}}$, the restore it pays roughly offsetting the prefill it saves; the decisive win is the fast path, which removes the decode entirely. The bars are computed from the measured prefill cost $\tau_p \approx 0.843$\,ms/token (HW-1) with the measured decode and the derived scan and restore constants (HW-2). Coordinates are emitted by our committed simulator.}
\label{fig:latency-breakdown}
\end{figure}

Figure~\ref{fig:latency-breakdown} decomposes those latencies into their additive cost components at a representative entry length, from the same constants. The baseline $L_{\text{cold}}$ of Eq.~\eqref{eq:lcold} is dominated by the entry prefill plus one decode; the fast path $L_{\text{fast}}$ of Eq.~\eqref{eq:lfast} is the lone vocabulary scan, a small fraction of either term; and the fall-through $L_{\text{fall}}$ of Eq.~\eqref{eq:fallthrough} lies between its two endpoints, the exact-match resume at $\Delta = 0$ (a constant-time restore plus one decode, well below the baseline) and the full-residual worst case at $\Delta = L_{\text{entry}}$ (a residual prefill atop the restore), with typical resumes sitting near the $\Delta = 0$ bar. The decomposition makes the mechanism concrete: pre-positioning takes the prefill term off the served path, replacing it with nothing on the fast path and with a small restore on the typical fall-through, so the critical-path cost a request actually pays is the short bar rather than the tall one. Like the other curves here these bars are computed from the measured prefill, decode, scan, and restore constants (HW-1, HW-2).

\begin{figure}[htbp]
\centering
\begin{subfigure}{0.48\textwidth}
\centering
\begin{tikzpicture}
\begin{axis}[
    width=\linewidth, height=5cm,
    xmode=log,
    xlabel={Query-to-update ratio $\rho = \lambda_q/\lambda_u$},
    ylabel={Fraction},
    xmin=0.1, xmax=10, ymin=0, ymax=1,
    legend pos=north west, legend cell align=left,
    grid=major,
    tick label style={font=\scriptsize},
    label style={font=\small},
    legend style={font=\scriptsize}
]
\addplot[blue!70!black, very thick, mark=none] coordinates {(0.1,0.0909) (0.2,0.1667) (0.3,0.2308) (0.5,0.3333) (0.7,0.4118) (1,0.5) (1.5,0.6) (2,0.6667) (3,0.75) (5,0.8333) (7,0.875) (10,0.9091)};
\addlegendentry{$h$ (hit rate)}
\addplot[orange!85!black, very thick, dashed, mark=none] coordinates {(0.1,0.9091) (0.2,0.8333) (0.3,0.7692) (0.5,0.6667) (0.7,0.5882) (1,0.5) (1.5,0.4) (2,0.3333) (3,0.25) (5,0.1667) (7,0.125) (10,0.0909)};
\addlegendentry{$w = 1-h$ (wasted)}
\addplot[black, only marks, mark=*, mark size=1.8pt] coordinates {(0.2,0.125) (0.5,0.3047) (1,0.4844) (2,0.6641) (3,0.6875) (5,0.7812)};
\addlegendentry{$h$ measured}
\end{axis}
\end{tikzpicture}
\caption{Hit rate and wasted-work fraction.}
\end{subfigure}
\hfill
\begin{subfigure}{0.48\textwidth}
\centering
\begin{tikzpicture}
\begin{axis}[
    width=\linewidth, height=5cm,
    xmode=log,
    xlabel={Query-to-update ratio $\rho = \lambda_q/\lambda_u$},
    ylabel={Effective first-token latency (ms)},
    xmin=0.1, xmax=10, ymin=0, ymax=60,
    legend pos=south west, legend cell align=left,
    grid=major,
    tick label style={font=\scriptsize},
    label style={font=\small},
    legend style={font=\scriptsize}
]
\addplot[blue!70!black, very thick, mark=none] coordinates {(0.1,49.712) (0.2,45.897) (0.3,42.668) (0.5,37.503) (0.7,33.553) (1,29.109) (1.5,24.073) (2,20.715) (3,16.518) (5,12.321) (7,10.223) (10,8.506)};
\addlegendentry{$\mathbb{E}[L_{\text{pp}}]$}
\addplot[orange!85!black, very thick, dashed, mark=none] coordinates {(0.1,54.29) (10,54.29)};
\addlegendentry{$L_{\text{cold}}$ (baseline)}
\end{axis}
\end{tikzpicture}
\caption{Effective first-token latency.}
\end{subfigure}
\caption{Amortized effect of idle-window pre-positioning as the query-to-update ratio $\rho = \lambda_q/\lambda_u$ varies, from the closed-form session-trace model of Section~\ref{sec:problem}. (a) The per-pre-position hit rate $h = \rho/(1+\rho)$ of Eq.~\eqref{eq:hitrate} rises with $\rho$ while the wasted-work fraction $w = 1-h$ falls: update-heavy sessions (small $\rho$) waste most pre-positions and query-heavy sessions (large $\rho$) consume nearly all of them. (b) The effective first-token latency $\mathbb{E}[L_{\text{pp}}]$ of Eq.~\eqref{eq:expected-latency} stays below the baseline $L_{\text{cold}}$ for every $\rho > 0$, consistent with the net-benefit condition of Eq.~\eqref{eq:netbenefit}, and approaches the served latency as $\rho$ grows. The points in panel (a) are the measured per-pre-position hit rate from a live session trace (HW-4), each over $128$ pre-positions, which track the closed form and run modestly below it. Curves use a gate-fire probability $g = 0.9453$, the measured gate coverage at $\tau = 2.0$, the measured representative entry length of $18$ tokens, and the measured latency constants (HW-1, HW-2). Coordinates are emitted by our committed simulator.}
\label{fig:amortization}
\end{figure}
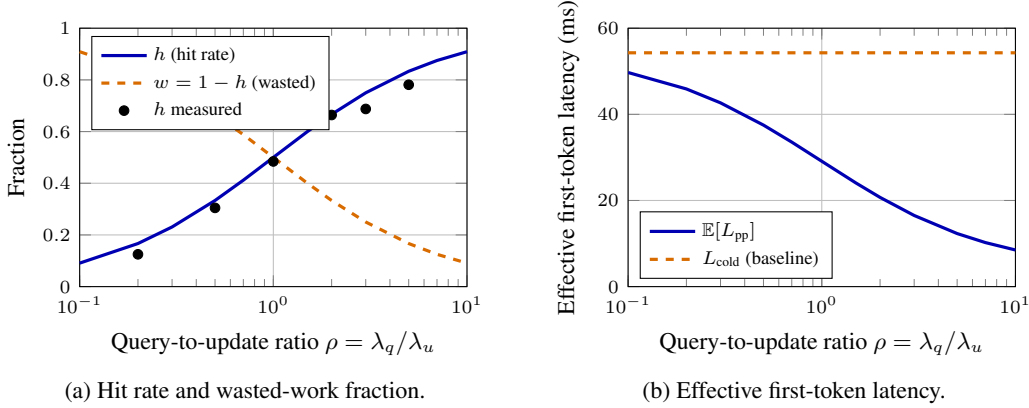

Figure~\ref{fig:amortization} turns to the cross-request amortization of Section~\ref{sec:problem}, plotting the closed-form session-trace model as the query-to-update ratio $\rho = \lambda_q/\lambda_u$ varies. The per-pre-position hit rate $h = \rho/(1+\rho)$ of Eq.~\eqref{eq:hitrate} and the complementary wasted-work fraction $w = 1-h$ quantify how often an idle pre-position is consumed before the next state mutation invalidates it: update-heavy sessions waste most pre-positions, whereas query-heavy sessions consume nearly all of them. The effective first-token latency $\mathbb{E}[L_{\text{pp}}]$ of Eq.~\eqref{eq:expected-latency} falls monotonically toward the served latency as $\rho$ grows and stays below the baseline $L_{\text{cold}}$ for every positive hit rate, in agreement with the net-benefit condition of Eq.~\eqref{eq:netbenefit}: because the entry prefill is pre-paid in the idle window, even a fall-through is cheaper than a cold entry, so the wasted work bears only on energy and never on the latency of a served query. The measured per-pre-position hit rate from a live session trace (HW-4) is overlaid on panel (a) and tracks the i.i.d. closed form, running modestly below it across the range, each point resting on $128$ pre-positions with a Wilson $95\%$ interval. The session-trace model is computed by our committed simulator from the measured latency constants (HW-1, HW-2) under a gate-fire probability anchored to the measured gate coverage at $\tau = 2.0$ (HW-3).

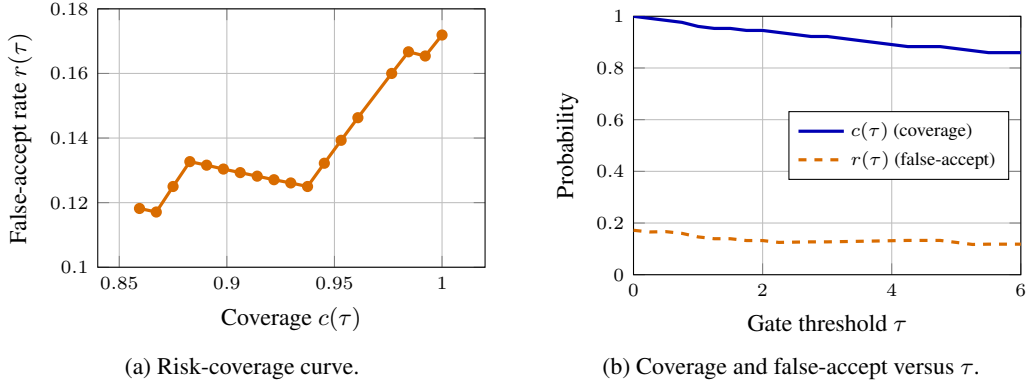
\begin{figure}[htbp]
\centering
\begin{subfigure}{0.48\textwidth}
\centering
\begin{tikzpicture}
\begin{axis}[
    width=\linewidth, height=5cm,
    xlabel={Coverage $c(\tau)$},
    ylabel={False-accept rate $r(\tau)$},
    xmin=0.84, xmax=1.02, ymin=0.1, ymax=0.18,
    grid=major,
    tick label style={font=\scriptsize},
    label style={font=\small}
]
\addplot[orange!85!black, very thick, mark=*, mark size=1.5pt] coordinates {(0.8594,0.1182) (0.8672,0.1171) (0.875,0.125) (0.8828,0.1327) (0.8906,0.1316) (0.8984,0.1304) (0.9062,0.1293) (0.9141,0.1282) (0.9219,0.1271) (0.9297,0.1261) (0.9375,0.125) (0.9453,0.1322) (0.9531,0.1393) (0.9609,0.1463) (0.9766,0.16) (0.9844,0.1667) (0.9922,0.1654) (1,0.1719)};
\end{axis}
\end{tikzpicture}
\caption{Risk-coverage curve.}
\end{subfigure}
\hfill
\begin{subfigure}{0.48\textwidth}
\centering
\begin{tikzpicture}
\begin{axis}[
    width=\linewidth, height=5cm,
    xlabel={Gate threshold $\tau$},
    ylabel={Probability},
    xmin=0, xmax=6, ymin=0, ymax=1,
    legend cell align=left,
    grid=major,
    tick label style={font=\scriptsize},
    label style={font=\small},
    legend style={at={(0.97,0.5)}, anchor=east, font=\scriptsize}
]
\addplot[blue!70!black, very thick, mark=none] coordinates {(0,1) (0.25,0.9922) (0.5,0.9844) (0.75,0.9766) (1,0.9609) (1.25,0.9531) (1.5,0.9531) (1.75,0.9453) (2,0.9453) (2.25,0.9375) (2.5,0.9297) (2.75,0.9219) (3,0.9219) (3.25,0.9141) (3.5,0.9062) (3.75,0.8984) (4,0.8906) (4.25,0.8828) (4.5,0.8828) (4.75,0.8828) (5,0.875) (5.25,0.8672) (5.5,0.8594) (5.75,0.8594) (6,0.8594)};
\addlegendentry{$c(\tau)$ (coverage)}
\addplot[orange!85!black, very thick, dashed, mark=none] coordinates {(0,0.1719) (0.25,0.1654) (0.5,0.1667) (0.75,0.16) (1,0.1463) (1.25,0.1393) (1.5,0.1393) (1.75,0.1322) (2,0.1322) (2.25,0.125) (2.5,0.1261) (2.75,0.1271) (3,0.1271) (3.25,0.1282) (3.5,0.1293) (3.75,0.1304) (4,0.1316) (4.25,0.1327) (4.5,0.1327) (4.75,0.1327) (5,0.125) (5.25,0.1171) (5.5,0.1182) (5.75,0.1182) (6,0.1182)};
\addlegendentry{$r(\tau)$ (false-accept)}
\end{axis}
\end{tikzpicture}
\caption{Coverage and false-accept versus $\tau$.}
\end{subfigure}
\caption{Selective-prediction behavior of the confidence gate as the threshold $\tau$ sweeps, from the measured (logit-gap, correctness) pairs of the host run ($128$ independent decision points, $121$ served at $\tau = 2.0$, fast-answer set $|\mathcal{A}| = 31$). (a) The risk-coverage curve traces the false-accept rate $r(\tau)$ of Eq.~\eqref{eq:risk} against the coverage $c(\tau)$ of Eq.~\eqref{eq:coverage} as $\tau$ varies: serving more requests (higher coverage) generally admits more false accepts, and the gate of Eq.~\eqref{eq:gate} picks an operating point on this curve. (b) Coverage is monotone non-increasing in $\tau$ by construction; the false-accept rate tends to fall but is not monotone, staying within a narrow $0.12$ to $0.17$ band. At the coded threshold $\tau = 2.0$ the gate covers $0.945$ of decision points at a $0.132$ false-accept rate (Wilson $95\%$ intervals $[0.89, 0.97]$ and $[0.083, 0.204]$), agreeing with the full path about $87\%$ of the time when it fires. These curves are measured: the gate fires with the 70B-class model (the gaps separate correct from incorrect predictions) and never fires with the 8B model (Section~\ref{sec:applicability}). Coordinates are emitted by our committed simulator.}
\label{fig:risk-coverage}
\end{figure}

Figure~\ref{fig:risk-coverage} characterizes the confidence gate of Section~\ref{sec:gate-formulation} as a selective-prediction tradeoff, from the measured (logit-gap, correctness) pairs of the host run (HW-3): $128$ independent decision points, one canonical query per non-overlapping data window, with $121$ served at $\tau = 2.0$ and a fast-answer set of $|\mathcal{A}| = 31$ tokens. As the threshold $\tau$ of the gate of Eq.~\eqref{eq:gate} sweeps, the coverage $c(\tau)$ of Eq.~\eqref{eq:coverage} is monotone non-increasing by construction; the false-accept rate $r(\tau)$ of Eq.~\eqref{eq:risk} tends to fall as the threshold rises but is not monotone, since raising $\tau$ removes borderline accepts that may have been correct or incorrect, and most false accepts here stem from genuine model uncertainty rather than a small gap (below). The measured curve runs from coverage $1.0$ at a $0.17$ false-accept rate (low $\tau$) down to coverage $0.86$ at a $0.12$ false-accept rate (high $\tau$), staying within a narrow $0.12$ to $0.17$ false-accept band throughout. At the coded threshold $\tau = 2.0$ the gate covers $0.945$ of decision points (Wilson $95\%$ interval $[0.89, 0.97]$) at a $0.132$ false-accept rate ($[0.083, 0.204]$), so it agrees with the model's own full-path output about $87\%$ of the time when it fires. The divergence decomposes cleanly: of the $16$ false accepts among the $121$ served, $12$ fall on windows where the model's own full path is itself unsure (a genuinely ambiguous regime) and only $4$ are pre-commitment divergence, where the question-independent token disagrees with a full path that was right. The pre-positioning trade of Section~\ref{sec:fastpath}, then, accounts for a small minority of the cost; the rest is the model's own uncertainty, which a full decode would also incur. That the gate fires at all is the model-capability observation of Section~\ref{sec:applicability}: with the 70B-class model the gaps separate correct from incorrect predictions, whereas the 8B model never clears the threshold.

\subsection{Model Capability and Single-GPU Operation}
\label{sec:applicability}
Two measured findings frame when speculative pre-positioning applies, on the single binary market-trend task and the two models we measure. The first is a model-capability observation. On this task the confidence gate fires only for a model that reads it decisively. The 70B-class target model does: its decision-point logit gaps span a wide range, from below $2$ to about $20$, it answers the trend correctly, and coverage is $1.0$ at the low end of the threshold sweep (Figure~\ref{fig:risk-coverage}). The 8B model (BF16) does not: its decision-point gap is about $0.65$ and direction-insensitive (it reports a downward trend on clean upward data), so it never clears the threshold and coverage is zero. The fast path is therefore a no-op for the 8B model, and we anchor the headline evaluation to the 70B-class model, keeping the 8B as the contrast that motivates the observation: on this task pre-positioning pays only when the model is confident, exactly the regime in which a cached single-token answer is trustworthy. Whether the transition is a sharp threshold or graded across model scales and tasks is left to future work, since two models on one task cannot settle it. The second finding concerns deployment. The 70B-class model at 4-bit precision fits one H100, so distributing it across two accelerators is a small regression rather than a gain: the single decode step and the cold first token both rise modestly with no change in gate behavior, the extra accelerator adding cross-device communication without relieving a memory bottleneck that does not exist at this size. We therefore report the single-accelerator configuration, anchored at a single decode step of $39.12$\,ms and a cold first token of $53.1$\,ms (HW-2, HW-5).

\subsection{Measured Systems Quantities}
\label{sec:measured}
Table~\ref{tab:measured} collects the measured systems quantities behind the evaluation, all from the host run (single H100, a 70B-class target model at 4-bit precision, TP$=1$). The end-to-end first-token latency arms (HW-5) are the headline: a cold first token of $53.1$\,ms (P50) and $66.9$\,ms (P99) over $144$ observations, a fall-through of $53.0$\,ms (P50), and a fast-path first token of $1.01$\,ms (P50), $1.35$\,ms (P99), mean $1.04 \pm 0.52$\,ms over $128$ independent gate-fires. The fall-through and the cold path nearly coincide here because the measured entry is short, so the entry prefill it saves is comparable to the state restore it pays; the decisive win is the fast path. That speedup, more than fiftyfold, is against the stateful cold path, which already caches the accumulated context and pays only the entry prefill and one decode. The contribution is sharper against the relevant prior art, a prefix or key-value cache: it removes the entry prefill but still runs one decode step, the measured $T_{\text{decode}} = 39.12$\,ms (HW-2), to produce the first token, because cached tensors expose the state but not the next token. The ready-distribution cache instead moves that decode into the idle window and caches its output, so the served path runs no decode at all and returns the first token in $1.01$\,ms, a nearly fortyfold reduction; the rest of the margin to the cold figure is the pre-paid entry prefill, and the decode, at $74\%$ of the cold path, is the dominant term either comparison removes. The marginal vocabulary scan is the derived $0.664$\,ms and the state restore the derived $14.47$\,ms (HW-2); the gate calibration (HW-3) and the hit rate (HW-4) are plotted in Figures~\ref{fig:risk-coverage} and \ref{fig:amortization}. Each pre-position draws about $63.8$\,J of marginal board energy (HW-6), the only cost it adds under the amortization condition (Section~\ref{sec:discussion}).

\begin{table}[htbp]
\caption{Measured systems quantities (single H100, a 70B-class target model at 4-bit precision, TP$=1$, 2026-06-28). Each quantity is measured or a labeled residual of measured quantities: the prefill and decode constants from a microbenchmark, the gate calibration from measured (logit-gap, correctness) pairs over $128$ decision points, the hit rate from a live session trace, the end-to-end first-token latency with and without pre-positioning, and the idle-time board energy. The fast-path scan and the state restore are algebraic residuals, not standalone microbenchmarks: scan $= L_{\text{fast}}^{\text{p50}}$ minus the net per-request floor, and $T_{\text{restore}} = L_{\text{fall}}^{\text{p50}} - T_{\text{decode}}$. The fast-path arm is the headline, more than fiftyfold below the cold first token.}
\label{tab:measured}
\centering
\small
\begin{tabular}{llll}
\toprule
Quantity & Symbol & Measured value & Source \\
\midrule
Prefill cost per token & $\tau_p$ & $0.843$\,ms/tok & microbench (HW-1) \\
Single-step decode time & $T_{\text{decode}}$ & $39.12$\,ms & microbench (HW-2) \\
Fast-path scan & $L_{\text{fast}}$ & $1.01$\,ms (scan $0.664$) & derived (HW-2) \\
State-restore cost & $T_{\text{restore}}$ & $14.47$\,ms & derived (HW-2) \\
Gate at $\tau = 2.0$ & $c,\, r$ & cover.\ $0.945$, f.a.\ $0.132$ & measured pairs (HW-3) \\
Hit rate & $h(\rho)$ & $0.13$ to $0.78$ & live trace (HW-4) \\
First-token latency & P50/P99 & cold $53.1/66.9$; fast $1.01/1.35$ & live session (HW-5) \\
Energy per pre-position & $E_{\text{pp}}$ & $63.8$\,J ($191$\,W wasted at $\rho{=}1$) & measured (HW-6) \\
\bottomrule
\end{tabular}
\end{table}

\section{Discussion}
\label{sec:discussion}

Under the amortization condition of Eq.~\eqref{eq:amortization} the pre-position pass runs in otherwise-idle time, adding nothing to wall-clock latency; it costs two things instead. The first is energy: a pre-position invalidated before use is wasted work, the fraction $w = 1 - h$ of Eq.~\eqref{eq:hitrate}, drawing power, not latency. We measure one pre-position at about $63.8$\,J of marginal board energy (HW-6, board power rising from $117.6$\,W idle to $500.1$\,W under a back-to-back pre-position loop); at the balanced ratio $\rho = 1$, where half of all pre-positions go unused, the wasted share averages about $191$\,W of board power, the energy paid to keep the next token waiting. The second is the rare false accept: the fast path serves a cached token unverified, so a miscalibrated gate can serve one the full path would not, at the rate $r(\tau)$ of Eq.~\eqref{eq:risk}. Both are bounded and tunable, by $\tau$ and the query-to-update ratio.

Pre-positioning pays in the query-heavy regime: at large $\rho$ (Figure~\ref{fig:amortization}) nearly every pre-position is consumed, whereas an update-heavy session wastes most for idle energy. Best-effort at the lowest priority, it bounds that worst case: every miss takes the baseline path of Eq.~\eqref{eq:lcold} or the fall-through of Eq.~\eqref{eq:fallthrough}. It thus composes with, not replaces, what a stateful system runs, yielding to any arriving request and adding only a vocabulary-sized vector beside the key-value cache; a missed request still benefits from the pre-paid entry prefill, under the invalidation discipline of Section~\ref{sec:invalidation}.

\subsection{Limitations}
Several conditions bound the mechanism. First, the fast path's safety rests entirely on gate calibration: with no verification on the served token, a miscalibrated threshold raises the false-accept rate of Eq.~\eqref{eq:risk} directly, so the calibration (HW-3) is load-bearing, and the measured false-accept rate of about $0.13$ at the coded threshold is a real cost the fast-answer set must keep bounded. Second, the benefit needs a genuinely idle window: if Eq.~\eqref{eq:amortization} fails because it is saturated, the lowest-priority guard caps the downside to wasted energy, not a speedup. Third, the single-token fast path serves only a domain-scoped class, the fast-answer set $\mathcal{A}$ of Eq.~\eqref{eq:gate}; queries outside it fall through to the still-cheaper normal path. Fourth, the fast path requires a model confident on the domain task: the gate fires for the 70B-class model and never for the 8B model (Section~\ref{sec:applicability}), so on a task the model reads indecisively the mechanism is a no-op, not a regression. Fifth, the 70B-class model we measure runs at 4-bit precision, which is what fits a single accelerator and exercises the gate, whereas the 8B reference is BF16; the headline therefore departs from a strict same-precision comparison, and a BF16 70B-class re-run (which needs two accelerators) is future work. Finally, it depends on persistent session state with no analogue in a stateless request loop, an optimization for the companions' stateful-session setting \cite{norgren2026layerscale,norgren2026toolcall}, not a general-purpose decoding technique.

\section{Related Work}
\label{sec:related}

Speculative decoding speculates tokens \emph{within} a generation with a draft model and verification \cite{leviathan2023fast,chen2023spec}, and prompt-lookup decoding does so draft-free but still verified \cite{saxena2023pld}; we instead speculate the decode \emph{entry point across the request boundary} and the \emph{output distribution}, with no fast-path verification, the gate a calibrated confidence test rather than a draft-versus-target match.

Prefix and key-value caches \cite{kwon2023vllm,zheng2024sglang}, attention-state reuse \cite{gim2024promptcache}, and continuous batching \cite{yu2022orca} reuse stored state across requests but keep the key-value tensors alone, never the output distribution computed ahead of the request; our ready-distribution cache caches the answer at the decision point, collapsing the critical path to one vocabulary scan. Early exit \cite{schuster2022calm} and model cascades \cite{chen2023frugalgpt} spend confidence-gated compute live on the critical path; ours, framed by selective prediction \cite{elyaniv2010selective}, consumes a pre-positioned distribution and abstains by falling through. All of it builds on our streaming \cite{norgren2026layerscale} and tool-calling \cite{norgren2026toolcall} companions.

\section{Conclusion}
\label{sec:conclusion}

A stateful session spends part of its life idle: between a data update and the next query, or between a tool call and its result. We have argued that this idle time, not the next request's critical path, is where the cost of the next decode entry point belongs. Speculative pre-positioning decodes the session forward to its next decision point during those windows with the target model's own forward pass and no draft model, so the next request either resumes delta-only or is answered directly from a cached output distribution. We formalized the decision point, the amortization condition, and the net-benefit condition (Section~\ref{sec:problem}); introduced the ready-distribution cache, which retains the output distribution at the decision point and not only the key-value state (Section~\ref{sec:ready-cache}); built a confidence-gated single-token fast path that serves one cached token when the top-two-logit gap clears a calibrated threshold and otherwise falls through to a still-cheaper path (Section~\ref{sec:fastpath}); and showed one mechanism covers both streaming and agentic sessions (Section~\ref{sec:agentic}), under an invalidation discipline that never serves a stale prediction (Section~\ref{sec:invalidation}).

The analytical model gives a fast-path speedup $S$ (Eq.~\eqref{eq:fastspeedup}) that grows with the pre-positioned entry length (Figure~\ref{fig:speedup}), bounded in risk by the gate threshold along the measured selective-prediction curve of Figure~\ref{fig:risk-coverage}. On a host run the mechanism delivers a measured end-to-end first token of $1.01$\,ms, removing from the critical path the one decode step a prefix cache must still run (about $39$\,ms), a nearly fortyfold reduction and more than fiftyfold against the stateful cold path (HW-5, Table~\ref{tab:measured}); but this holds only for a model confident on the domain task: the gate fires for a 70B-class target model and never for an 8B model, so on the task and models we measure speculative pre-positioning is an optimization for a capable model on a decisive task, not a universal one. The construction extends the stateful-session foundation of our streaming \cite{norgren2026layerscale} and tool-calling \cite{norgren2026toolcall} companions, putting the session's persistent state to a new use: turning the idle time between requests into the place where the next request's first token is already waiting.

\bibliographystyle{plainnat}

\end{document}